\documentclass[]{fairmeta}
\usepackage{xspace}
\usepackage[table]{xcolor}
\usepackage{overpic}
\usepackage{tikz}
\usepackage{amsmath}
\usepackage{amssymb}
\usepackage{mathtools}
\usepackage{amsthm}
\usepackage{bm}

\newcommand\our{{\fontfamily{lmtt}\selectfont RepFusion}\xspace}
\newcommand{\thinparagraph}[1]{\vspace{0.15em}\noindent\textbf{#1}\enspace}

\newcommand{\flopsbreakdown}[4]{%
    \begin{tikzpicture}[baseline=-0.5ex, x=1cm, y=1ex]
        \path[use as bounding box] (-2.05,-1.0) rectangle (2.05,1.0);
        \draw[black!45, line width=0.35pt] (0,-0.75) -- (0,0.75);
        \draw[fill=blue!55, draw=none] (-#1,-0.45) rectangle (0,0.45);
        \draw[fill=orange!75, draw=none] (0,-0.45) rectangle (#2,0.45);
        \node[anchor=east, inner sep=1pt, font=\scriptsize\ttfamily, text=black!70] at (-1.5,0) {#3};
        \node[anchor=west, inner sep=1pt, font=\scriptsize\ttfamily, text=black!70] at (1.5,0) {#4};
    \end{tikzpicture}%
}

\title{\our: Leveraging Multimodal Priors for Denoising in Representation Space}

\author[1,2]{Xichen Pan}
\author[1]{Aashu Singh}
\author[1,*]{Satya Narayan Shukla}
\author[1]{Xiangjun Fan}
\author[1,\dagger]{Shlok Kumar Mishra}
\author[2,\dagger]{Saining Xie}
\renewcommand\authorlist{%
  \authorformat[1,2]{Xichen Pan},
  \authorformat[1]{Aashu Singh},
  \authorformat[1,*]{Satya Narayan Shukla},
  \authorformat[1]{Xiangjun Fan},
  \authorformat[1,\dagger]{Shlok Kumar Mishra},\\
  \authorformat[2,\dagger]{Saining Xie}
}
\affiliation[1]{Meta AI}
\affiliation[2]{New York University}

\contribution[*]{Work done at Meta}
\contribution[\dagger]{Equal advising}

\abstract{Large language models (LLMs) are widely used in text-to-image (T2I) systems, but they are typically limited to text encoding, while denoising is handled by newly trained generative backbones. The emergence of representation autoencoders (RAEs) shifts the generation target toward semantically structured visual representations, creating a latent space that is more compatible with pretrained LLM priors. Inspired by multimodal LLMs (MLLMs), where an MLP projector is sufficient to align clean visual representations with a pretrained LLM, we repurpose the MLLM itself as a noisy representation encoder, extending this mechanism from clean to noisy inputs. We present \our, which uses the resulting MLLM outputs as the conditioning signal for a diffusion transformer. In controlled comparisons at similar inference budgets, \our outperforms baselines that devote comparable capacity to newly initialized denoisers. Reusing MLLM outputs across denoising steps further accelerates inference while retaining performance comparable to per-step MLLM recomputation. These results demonstrate that MLLMs provide strong priors for denoising visual representations.
}

\date{\today}
\correspondence{Xichen Pan at \email{xichenpan@nyu.edu}}
\metadata[Project Page]{\url{https://xichenpan.com/repfusion}}

\begin{document}

\maketitle

\section{Introduction}

Text-to-image (T2I) generation is commonly formulated as conditional image generation, where image generators are conditioned on the outputs of text encoders. Alongside the evolution of image generators from GANs~\citep{gan} to diffusion models~\citep{ddpm}, text encoders have also progressed from LSTMs~\citep{lstm} to CLIP~\citep{clip} and T5~\citep{t5}. Recently, many systems have replaced these encoders with large language models (LLMs)~\citep{gpt, llama, llama3} due to their stronger representational capacity, richer world knowledge, in-context learning ability, and compatibility with unified multimodal models~\citep{metaquery}. However, in recent pipelines~\citep{pixart, luminanext, sana, qwenimage, flux2, zimage}, LLMs still primarily act as static text encoders that produce text embeddings, while diffusion transformers (DiTs)~\citep{dit} carry out the denoising trajectory and image synthesis.

This division of labor made sense in the VAE~\citep{vae} era. Diffusion models typically denoise VAE latents, and these latents were never designed to be ``read'' by pretrained language priors. They are low-dimensional, local, and optimized for reconstruction rather than semantics. As a result, even if one aims to bring an LLM closer to the denoising loop, it is unclear what the LLM should consume or why doing so would be beneficial.

\begin{figure*}[!t]
    \centering
    \begin{minipage}[t]{0.49\textwidth}
        \centering
        \includegraphics[width=\linewidth]{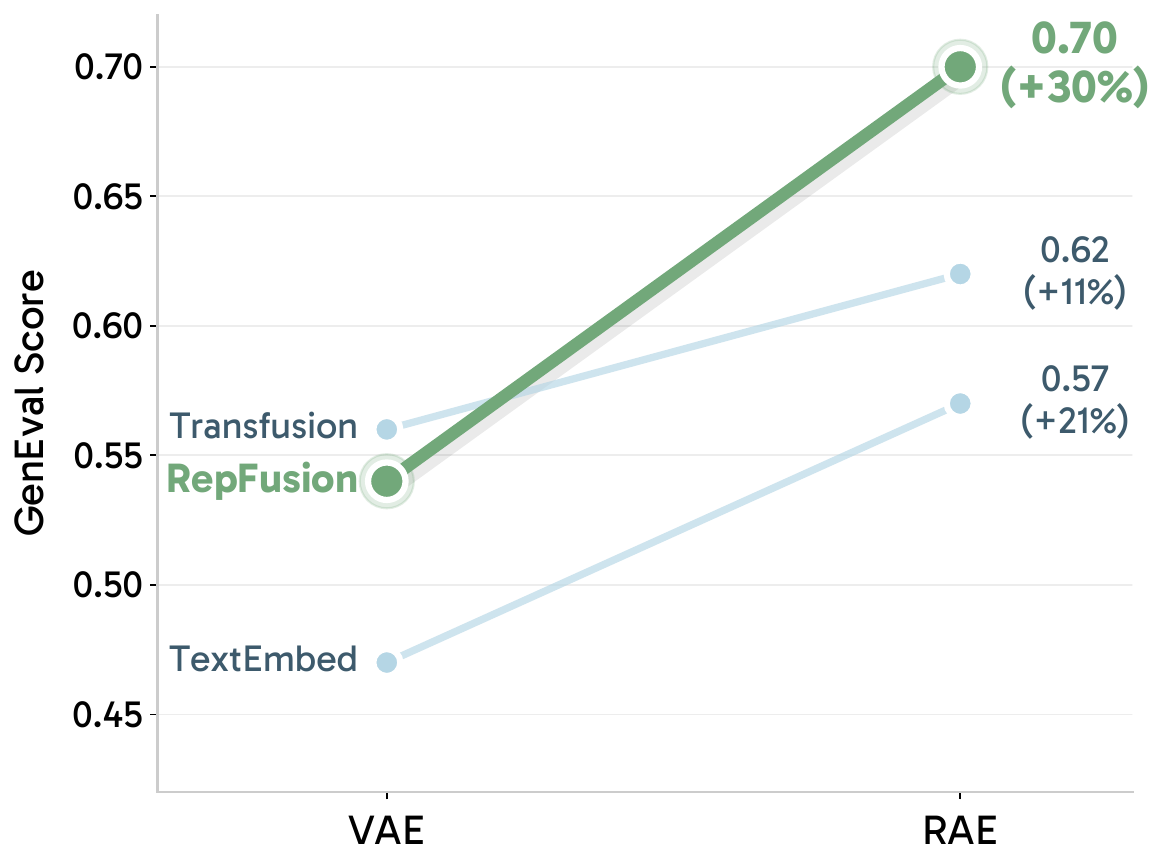}
        \caption{\mbox{GenEval} comparison when switching from VAEs to RAEs for three conditioning strategies: \mbox{TextEmbed} (conditioning a DiT with an LLM's last-layer text token embeddings following recent T2I practice~\citep{sana, qwenimage, flux2, zimage}), \mbox{Transfusion}~\citep{transfusion}, and \mbox{\our}. All three variants in this comparison use a 7B LLM, \mbox{TextEmbed} and \mbox{\our} also use a 1.3B DiT. \mbox{\our} feeds noisy visual representations into a pretrained MLLM and uses the resulting outputs to condition a DiT. It benefits most significantly from the transition, achieving a 30\% relative gain (+0.16 absolute improvement), compared to 21\% (+0.10) for \mbox{TextEmbed} and 11\% (+0.06) for \mbox{Transfusion}.}
        \label{fig:motivation}
    \end{minipage}
    \hfill
    \begin{minipage}[t]{0.49\textwidth}
        \centering
        \includegraphics[width=\linewidth]{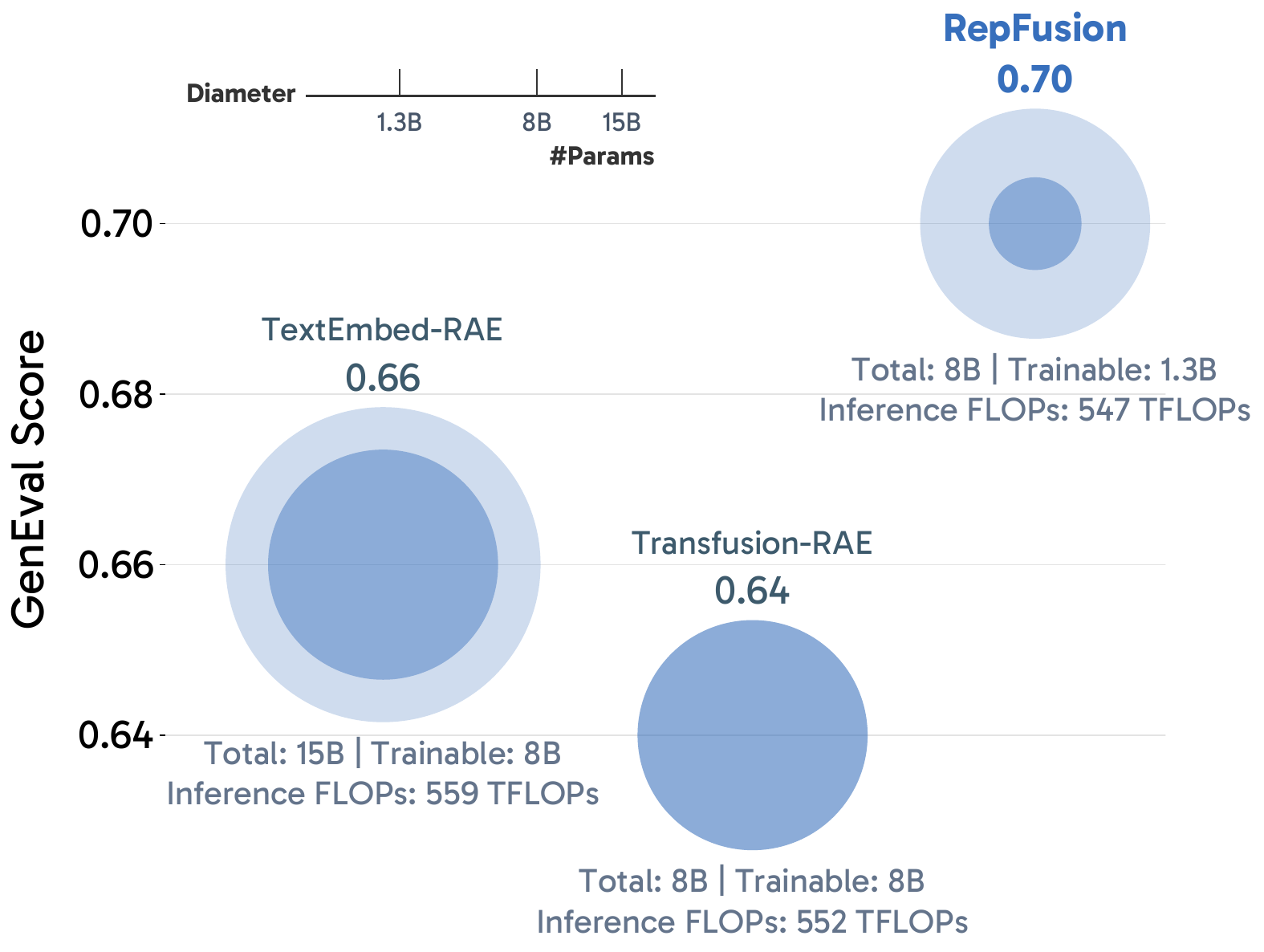}
        \caption{\mbox{GenEval} comparison across different conditioning strategies under similar inference FLOPs. Circle size denotes total parameters, and the inner disk denotes trainable parameters. Each method allocates roughly 8B parameters to modules that either process noisy visual latents or denoise them. \mbox{\our} fine-tunes only a 1.3B DiT and an MLP projector, yet outperforms \mbox{TextEmbed} and \mbox{Transfusion}, both of which train 8B parameters (a larger DiT and an LLM, respectively). This cross-method comparison suggests that MLLMs provide strong priors for denoising visual representations, and that repurposing them to encode noisy representations can be a more effective use of parameters than scaling newly initialized denoisers.}
        \label{fig:param_efficiency}
    \end{minipage}
\end{figure*}

Representation autoencoders (RAEs)~\citep{rae} change this picture. By moving generation from VAE latents to semantically structured visual representations, such as CLIP~\citep{clip} or DINO~\citep{dino} features, RAEs provide a denoising space that is both easier to optimize and more semantically meaningful. Furthermore, these developments bridge T2I and the feature spaces currently utilized by Multimodal LLMs (MLLMs).

In the multimodal understanding community, pretrained LLMs have demonstrated a simple yet powerful property: with an MLP projector, they can ingest clean visual representations and immediately become strong sequence models over multimodal tokens~\citep{llava}. This observation is usually discussed in the context of understanding and reasoning. Here, we take it as a design principle for generation: if an LLM can perceive clean visual representations, can it also process noisy counterparts during denoising?

Our answer is yes. As shown in Figure~\ref{fig:motivation}, the resulting system is highly effective and best suited to the RAE latent space. We present \our, a T2I model that treats a pretrained MLLM as a \emph{noisy representation encoder}. In addition to text inputs, we feed noisy RAE latents into an off-the-shelf MLLM by reusing its MLP projector. We keep the pretrained LLM backbone frozen and fine-tune only its projector. We then use the MLLM's output to condition a DiT that denoises in the same latent space. Conceptually, this design allows the pretrained MLLM to focus on what it does best: modeling structured visual representations.

This design first changes the capacity allocation picture beyond the standard ``make the denoiser bigger'' recipe. As shown in Figure~\ref{fig:param_efficiency}, under similar inference FLOPs, all compared systems allocate roughly 8B parameters to modules that either process noisy visual latents or denoise them: TextEmbed uses a 7B frozen MLLM text encoder and an 8B DiT, Transfusion uses an 8B joint denoising transformer, and \our uses the same 7B frozen MLLM together with a 1.3B DiT. We provide details on training the TextEmbed and Transfusion baselines in Appendix~\ref{app:baseline_training}. Despite fine-tuning only the DiT and an MLP projector, \our outperforms these baselines, showing that, across model families, allocating substantial model capacity to a frozen pretrained conditional encoder can outperform spending nearly the entire parameter budget on newly initialized denoising modules. This suggests that pretrained MLLMs carry priors that transfer beyond multimodal understanding: once the representation space is compatible, those priors can directly help denoise noisy visual representations.

We also compare against unified architectures such as Transfusion~\citep{transfusion}, which can be viewed as another way of exposing noisy visual information to language models. As shown in Figure~\ref{fig:motivation}, even when we upgrade such baselines to operate in the RAE latent space, the gains are smaller than those obtained by explicitly repurposing a frozen MLLM as a noisy encoder. In other words, moving from VAE to RAE helps, but by itself it does not unlock the full benefit of pretrained language priors.

During training, the MLLM and DiT in \our operate at the same timestep. Our default sampling procedure follows this design by running both models at every denoising step. To reduce inference cost, we examine whether the MLLM condition needs to be recomputed at every step. We uniformly divide the denoising steps into segments, running the MLLM once at the beginning of each segment and reusing its outputs while the DiT continues to run at every step. In our recomputation experiments, three MLLM updates over 30 denoising steps retain performance comparable to per-step recomputation, with a 2.4$\times$ sampling speedup.

In summary, this paper argues for a simple shift in perspective: many modern T2I systems already allocate substantial capacity to huge LLM text encoders, and RAEs provide a representation space where these encoders can do more than encode text. By letting frozen MLLMs take noisy visual representations as input, we obtain a strong and efficient prior for denoising in representation space. The main contributions are:
\begin{itemize}
    \item We show that frozen pretrained MLLMs can encode noisy RAE latents and provide useful denoising priors beyond static text conditioning.
    \item We demonstrate that allocating parameters to a frozen pretrained conditional encoder can outperform static text embedding baselines that spend comparable capacity on newly initialized denoisers.
    \item We show that the pretrained MLLM prior is strong: freezing it outperforms further jointly optimizing it for generation.
    \item We show that reusing MLLM outputs across denoising steps accelerates inference while retaining performance comparable to per-step MLLM recomputation.
\end{itemize}

\section{Related Work}

\thinparagraph{Text encoders in T2I}
Early conditional GANs use small text encoders such as LSTMs~\citep{lstm}, producing either global sentence embeddings~\citep{ganintcls, stackgan} or token-level embeddings~\citep{attngan}. Diffusion models later standardized text conditioning with frozen pretrained encoders that provide token embeddings for cross-attention. Stable Diffusion 1.5~\citep{sd1p5} popularized a CLIP~\citep{clip} text encoder. Recent systems increasingly scale the text encoder: Imagen~\citep{imagen} moves beyond CLIP to LLMs such as T5-XXL~\citep{t5}, and PixArt-$\alpha$~\citep{pixart}, Stable Diffusion 3~\citep{sd3}, and FLUX.1~\citep{flux} follow to ship with large T5-family encoders. Recent open-source models such as Lumina-Next~\citep{luminanext} and Sana~\citep{sana} adopt LLM encoders, and FLUX.2~\citep{flux2} further scales the LLM to a 24B-parameter Mistral Small 3~\citep{mistralsmall3}. Overall, modern T2I pipelines often devote billions of parameters to text encoders, motivating methods that better utilize their capacity.

\thinparagraph{From VAEs to RAEs}
Latent diffusion~\citep{sd1p5} popularized a key design choice in modern T2I models: instead of diffusing in pixel space, models denoise in the latent space of an autoencoder, making high-resolution generation tractable. Most systems adopt VAEs~\citep{vae} for this purpose, but VAE latents are heavily compressed and optimized for reconstruction, which limits their semantic expressiveness. RAEs~\citep{rae} avoid this bottleneck by pairing a decoder with a frozen pretrained encoder (e.g., CLIP~\citep{clip} or DINO~\citep{dino}), working with semantically rich latents that are easier to denoise. This shift removes the VAE bottleneck and brings T2I into representation spaces that pretrained MLLMs already handle well, creating a natural opportunity to leverage their priors beyond static text conditioning.

\thinparagraph{Integration of Language Models and Denoisers}
A growing line of work seeks tighter integration between conditional encoders and denoisers. Unified architectures such as Transfusion~\citep{transfusion} train a large transformer to jointly model language outputs and denoise VAE latents, aiming for a single modeling stack across modalities. Another direction builds compact interfaces between MLLMs and diffusion backbones, for example via learnable queries~\citep{metaquery, blip3o, scalerae} or joint attention~\citep{lmfusion, bagel}. In contrast, our focus is not on the conditioning mechanism, but on changing the content of the condition itself. We push MLLMs beyond text encoding, repurposing them to encode noisy representations and condition DiTs~\citep{dit}.

\begin{figure}[!t]
    \centering
    \includegraphics[width=0.55\linewidth]{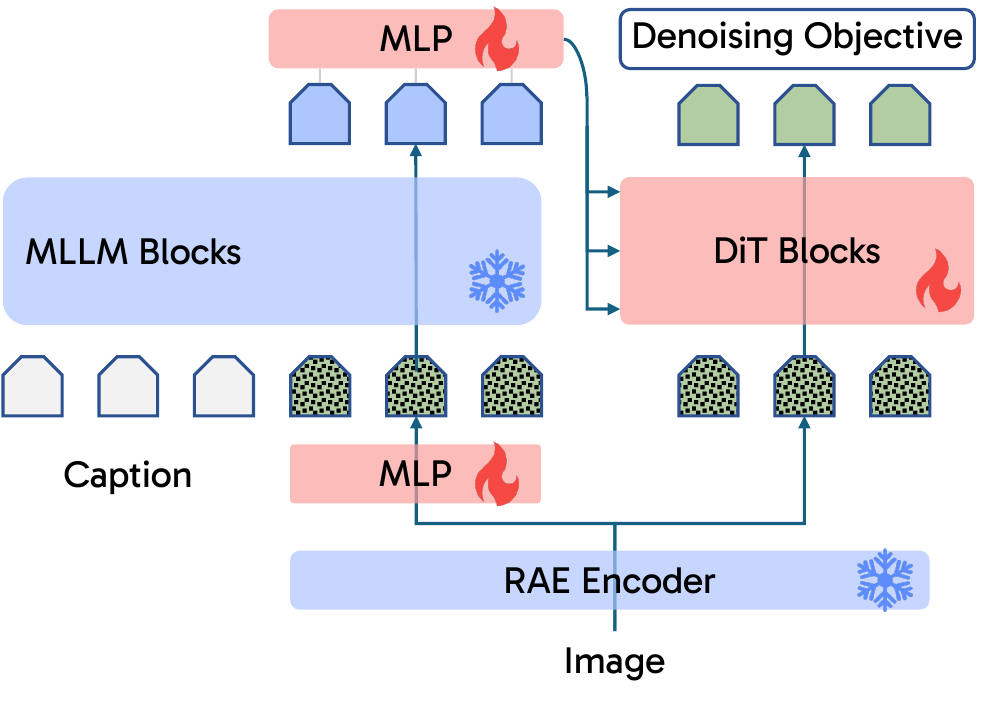}
    \caption{Overview of \our. Blue modules are frozen, while red modules are trainable. We reuse a pretrained MLLM to process the text prompt and noisy RAE latents. The noisy RAE latents are projected into the MLLM input space through an MLP projector, and the resulting outputs condition every DiT block via AdaLN modulation.}
    \label{fig:arch}
\end{figure}

\section{\our}
\label{sec:method}

This section first formalizes diffusion in visual representation space (Section~\ref{sec:preliminary}) and describes how \our uses an MLLM to encode noisy representations for DiT conditioning (Section~\ref{sec:method_design}). We then use controlled ablations to isolate the role of noisy representation inputs (Section~\ref{sec:noisy_representations}) and multimodal perception pretraining (Section~\ref{sec:perception_pretraining}). Finally, we break down how these ingredients improve over TextEmbed and Transfusion baselines (Section~\ref{sec:breakdown}). Unless otherwise specified, the variants discussed in this section use a 7B LLM backbone paired with a 1.3B DiT denoiser.

\subsection{Preliminary}
\label{sec:preliminary}
A flow matching T2I model is a conditional generative model. Given a text prompt $y$, we first obtain a text embedding $\bm{c}=E_\phi(y)$ with a typically frozen text encoder. The generative network is then conditioned on $\bm{c}$, either through cross attention~\citep{sd1p5} or adaptive normalization~\citep{dit}. In our setting, diffusion operates in a visual representation space: let $\bm{x}$ denote a clean visual representation, let $t$ be the timestep, and let $\bm{\epsilon}$ be the Gaussian noise. We adopt the $\bm{v}$-prediction parameterization~\citep{lipman2022flow, liu2022flow, albergo2022building}:
\begin{equation}
\bm{z}_t = t\,\bm{x} + (1-t)\,\bm{\epsilon}, \qquad
\bm{x}\sim p_{\text{data}}(\bm{x}).
\end{equation}
We follow the timestep shifting strategy of~\citet{rae, sd3}. For a base dimension $n$ and an effective data dimension $m$, the sampled timestep $t_n\sim\mathcal{U}(0,1)$ is shifted to $t = \frac{\alpha t_n}{1 + (\alpha - 1)t_n}$, where $\alpha = \sqrt{m/n}$.
Following~\citet{rae, sd3}, we use $n=4{,}096$ and set $m$ to the effective dimensionality of the visual representation; in our RAE setup, this gives $\alpha=12$.

The flow velocity is given by the time derivative of $\bm{z}_t$:
\begin{equation*}
\bm{v} \;=\; \bm{z}_t' \;=\; \bm{x} - \bm{\epsilon}.
\end{equation*}
We learn a conditional velocity field $\bm{v}_\theta(\bm{z}_t, t, \bm{c})$ by minimizing the standard flow matching objective~\citep{lipman2022flow, albergo2022building}:
\begin{equation*}
\mathcal{L}
:=
\mathbb{E}_{t, \bm{x}, \bm{\epsilon}}
\big\|
\bm{v}_\theta(\bm{z}_t, t, \bm{c}) - \bm{v}
\big\|^2,
\end{equation*}
where $\bm{v}_\theta$ is predicted by the diffusion model.

\subsection{Methods}
\label{sec:method_design}
In standard approaches, the conditioning $\bm{c}$ relies solely on the text $y$. In \our, as shown in Figure~\ref{fig:arch}, we augment the conditioning to also include the noisy visual representations, $\bm{z}_t$. This design allows the LLM to perceive the denoising trajectory.

Specifically, the LLM input consists of a sequence of text tokens followed by projected noisy visual tokens. Let $E_{\text{LLM}}$ denote the LLM, $P_\psi$ an MLP projector, and $\bm{e}_t$ the timestep embedding; we use the same notation for its projected forms in the visual representation space and the LLM hidden space. The conditioning $\bm{c}_t$ is defined as
\begin{equation}
    \bm{c}_t = \operatorname{Last}_N\!\left(E_{\text{LLM}}\left( [ y, P_\psi(\bm{z}_t + \bm{e}_t) ] \right)\right)
\label{eq:llm_output}
\end{equation}
where $[\cdot, \cdot]$ denotes sequence concatenation, and $\operatorname{Last}_N$ selects the final $N$ hidden states corresponding to the noisy visual tokens. We inject timestep information before the MLLM by adding the embedding $\bm{e}_t$ to $\bm{z}_t$ before applying the projection $P_\psi$, following \citet{transfusion}. During training, the MLLM condition and the DiT prediction are computed at the same sampled timestep $t$. Our default sampling procedure follows this design by running both models at every denoising step, recomputing $\bm{c}_t$ as $\bm{z}_t$ evolves. The LLM remains causal.

Following Decoupled Diffusion Transformer (DDT)~\citep{ddt}, we condition the DiT using adaptive layernorm~\citep{dit} without introducing additional cross-attention modules. Concretely, we adopt the AdaLN-Single variant from PixArt-$\alpha$~\citep{pixart}: a shared projection produces token-wise modulation parameters reused across blocks, and each transformer block adds a lightweight learned offset table $\bm{T}^{(\ell)} \in \mathbb{R}^{6 \times D}$ before splitting the resulting parameters into $(\bm{\beta}, \bm{\gamma}, \bm{\alpha})$ for the MSA and MLP branches.

Let $\bm{h}^{(\ell)}_t \in \mathbb{R}^{N \times D}$ denote the intermediate states at DiT block $\ell$ and timestep $t$, and let $\bm{c}_t \in \mathbb{R}^{N \times D_c}$ denote the selected LLM hidden states for the noisy visual tokens. In our setting, the token counts are aligned ($N = 576$). Following DiT~\citep{dit}, we inject the noise level by adding the timestep embedding and applying a SiLU nonlinearity:
\begin{equation*}
\tilde{\bm{c}}_t = \mathrm{SiLU}(\bm{c}_t + \bm{e}_t), \qquad \tilde{\bm{c}}_t \in \mathbb{R}^{N \times D_c}.
\end{equation*}
A shared linear layer predicts modulation parameters:
\begin{equation*}
\bm{m}_t = \mathrm{Linear}(\tilde{\bm{c}}_t) \in \mathbb{R}^{N \times 6D}.
\end{equation*}
For each block $\ell$, we interpret $\bm{m}_t$ as six token-wise $D$-dimensional modulation matrices and add a lightweight block-specific table $\bm{T}^{(\ell)} \in \mathbb{R}^{6 \times D}$ (broadcast over tokens) to obtain the final modulation parameters for the block. Splitting along the last channel dimension yields
\begin{equation*}
(\bm{\beta}^{(\ell)}_{t,\mathrm{msa}}, \bm{\gamma}^{(\ell)}_{t,\mathrm{msa}}, \bm{\alpha}^{(\ell)}_{t,\mathrm{msa}},
 \bm{\beta}^{(\ell)}_{t,\mathrm{mlp}}, \bm{\gamma}^{(\ell)}_{t,\mathrm{mlp}}, \bm{\alpha}^{(\ell)}_{t,\mathrm{mlp}}),
\end{equation*}
where each element lies in $\mathbb{R}^{N \times D}$.
Here, $\bm{\beta}$ and $\bm{\gamma}$ denote the shift and scale terms, respectively, and $\bm{\alpha}$ denotes the residual gate.

The modulation operator can be defined as $\mathrm{Mod}(\bm{u}; \bm{\gamma}, \bm{\beta}) = \bm{u} \odot (1+\bm{\gamma}) + \bm{\beta}$. Each block then computes:
\[
\begin{gathered}
\tilde{\bm{h}} = \mathrm{Mod}\!\left(\mathrm{RMSNorm}(\bm{h}^{(\ell)}_t);\ \bm{\gamma}^{(\ell)}_{t,\mathrm{msa}},\ \bm{\beta}^{(\ell)}_{t,\mathrm{msa}}\right), \\
\bm{h}'^{(\ell)}_t = \bm{h}^{(\ell)}_t + \bm{\alpha}^{(\ell)}_{t,\mathrm{msa}} \odot \mathrm{MSA}(\tilde{\bm{h}}), \\
\tilde{\bm{h}}' = \mathrm{Mod}\!\left(\mathrm{RMSNorm}(\bm{h}'^{(\ell)}_t);\ \bm{\gamma}^{(\ell)}_{t,\mathrm{mlp}},\ \bm{\beta}^{(\ell)}_{t,\mathrm{mlp}}\right), \\
\bm{h}^{(\ell+1)}_t = \bm{h}'^{(\ell)}_t + \bm{\alpha}^{(\ell)}_{t,\mathrm{mlp}} \odot \mathrm{MLP}(\tilde{\bm{h}}').
\end{gathered}
\]
Crucially, $\tilde{\bm{c}}_t$ is token-aligned with $\bm{h}^{(\ell)}_t$. As a result, the scale, shift, and residual gates are applied independently to each token rather than being broadcast over the sequence dimension.

\thinparagraph{MLLM Recomputation}
To reduce sampling cost, we uniformly divide the $S$ denoising steps into $U$ segments, running the MLLM once at the beginning of each segment and reusing its outputs within the segment. The DiT still runs at every denoising step. Unless otherwise specified, all \our results use the default setting $U=S$.

\begin{figure}[!t]
    \centering
    \makebox[\linewidth][c]{%
        \raisebox{1.5em}{\begin{minipage}[c]{0.23\linewidth}
            \centering
            \begin{tikzpicture}
                \node[
                    rounded corners=3pt,
                    draw=black!20,
                    fill=black!3,
                    inner sep=4pt,
                    text width=0.92\linewidth
                ] {
                    \centering\small\textbf{MetaQuery}\par
                    \vspace{0.35em}
                    \scriptsize
                    \begin{tabular}{@{}l@{}}
                        \textbf{Training FLOPs:} 35 T \\
                    \end{tabular}
                    \vspace{0.35em}
                    \par\noindent\colorbox{black!8}{%
                        \begin{minipage}{0.94\linewidth}
                            \centering
                            {\footnotesize\textbf{Matched Inference FLOPs}}\par
                            \vspace{0.5em}
                            \scriptsize
                            \begin{tabular}{@{}c@{\hspace{0.9em}}c@{}}
                                \textbf{FLOPs:} & 113 T $\rightarrow$ 552 T \\
                                \textbf{GenEval:} & 0.55 $\rightarrow$ 0.54
                            \end{tabular}
                        \end{minipage}}
                };
            \end{tikzpicture}
        \end{minipage}}\hfill
        \begin{overpic}[width=0.46\linewidth]{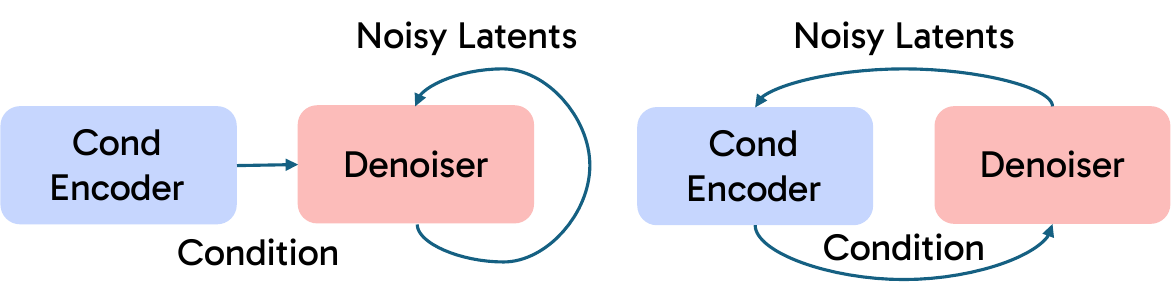}
            \put(13, -3){\small (a) MetaQuery}
            \put(63, -3){\small (b) \our}
        \end{overpic}\hfill
        \raisebox{1.5em}{\begin{minipage}[c]{0.23\linewidth}
            \centering
            \begin{tikzpicture}
                \node[
                    rounded corners=3pt,
                    draw=blue!25,
                    fill=blue!4,
                    inner sep=4pt,
                    text width=0.92\linewidth
                ] {
                    \centering\small\textbf{\our}\par
                    \vspace{0.35em}
                    \scriptsize
                    \begin{tabular}{@{}l@{}}
                        \textbf{Training FLOPs:} 35 T \\
                    \end{tabular}
                    \vspace{0.35em}
                    \par\noindent\colorbox{blue!8}{%
                        \begin{minipage}{0.94\linewidth}
                            \centering
                            {\footnotesize\textbf{Inference FLOPs}}\par
                            \vspace{0.5em}
                            \scriptsize
                            \begin{tabular}{@{}c@{\hspace{0.9em}}c@{}}
                                \textbf{FLOPs:} & 547 T \\
                                \textbf{GenEval:} & \textcolor{blue!65!black}{\textbf{0.70}}
                            \end{tabular}
                        \end{minipage}}
                };
            \end{tikzpicture}
        \end{minipage}}%
    }
\caption{High-level comparison between MetaQuery-style~\citep{metaquery} architectures (e.g., BLIP-3o~\citep{blip3o} and Scale-RAE~\citep{scalerae}) and \our. During training, both methods backpropagate gradients through the conditional encoder and denoiser, resulting in similar training FLOPs. Rerunning the MetaQuery conditional encoder with different timestep embeddings but no noisy visual inputs increases inference compute from 113 to 552 TFLOPs without improving GenEval (0.55 to 0.54). At a similar inference budget, \our reaches 0.70 with conditioning derived from noisy visual representations.}
\label{fig:metaquery}
\end{figure}

\begin{figure*}[!t]
    \centering
    \begin{subfigure}[t]{0.49\textwidth}
        \centering
        \includegraphics[width=\linewidth]{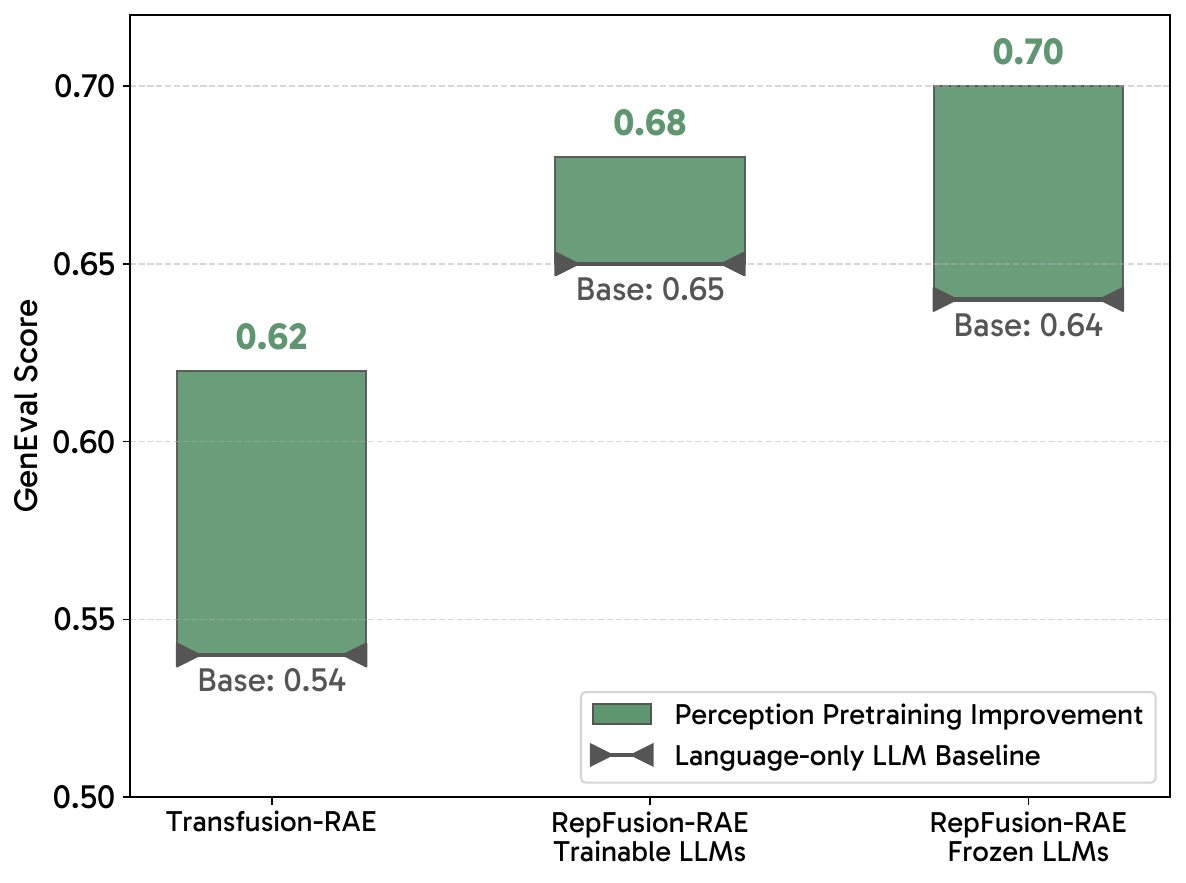}
        \caption{Effect of multimodal perception pretraining in the LLM backbone. Replacing an LLM with a perception-pretrained MLLM improves both Transfusion-RAE and \our under settings with frozen and trainable LLMs.}
        \label{fig:ptlm}
    \end{subfigure}
    \hfill
    \begin{subfigure}[t]{0.49\textwidth}
        \centering
        \includegraphics[width=\linewidth]{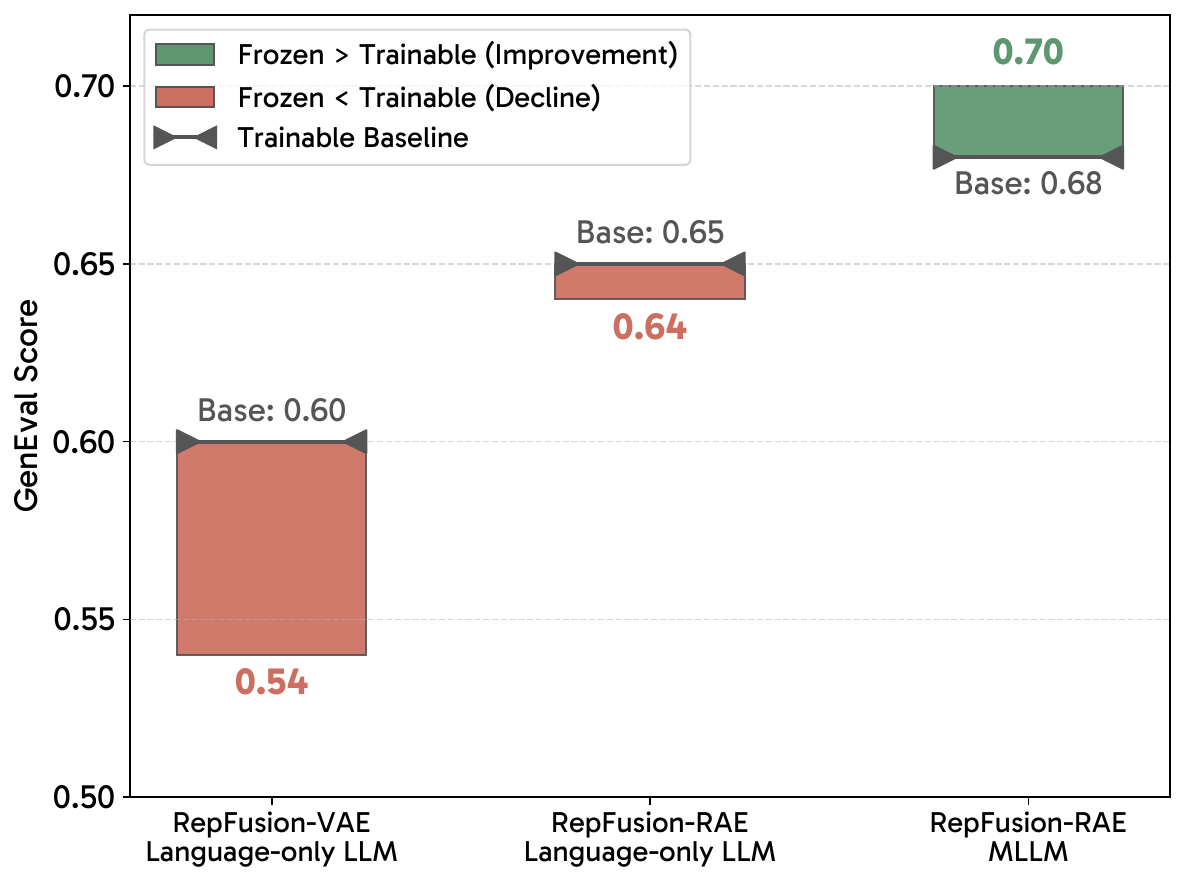}
        \caption{Effect of fine-tuning the LLM backbone. Fine-tuning helps when starting from a language-only LLM, but can hurt when starting from a perception-pretrained MLLM backbone in \our-RAE.}
        \label{fig:frozenllm}
    \end{subfigure}
    \caption{Ablations on multimodal perception pretraining and LLM fine-tuning. Perception-pretrained backbones consistently improve RAE diffusion, while fine-tuning benefits text-only backbones but may degrade performance when the backbone is already multimodally pretrained. This indicates that perception pretraining is a strong prior that can outperform joint optimization for generation.}
    \label{fig:ablation}
\end{figure*}

\subsection{Noisy Representation Input}
\label{sec:noisy_representations}
To assess the importance of conditioning the MLLM on noisy representations, as illustrated in Figure~\ref{fig:metaquery}, we construct a learnable query baseline by replacing the projected noisy RAE latents $P_{\psi}(\bm{z}_t+\bm{e}_t)$ in Equation~\ref{eq:llm_output} with $N$ learnable queries $\bm{Q}_{\eta}\in\mathbb{R}^{N\times D_c}$, following MetaQuery~\citep{metaquery}, while keeping the rest of the architecture unchanged. This baseline closely resembles BLIP-3o~\citep{blip3o} and Scale-RAE~\citep{scalerae}. With a frozen 7B MLLM and a 1.3B DiT, it reaches 0.55 on GenEval, whereas \our reaches 0.70.

Crucially, this gap is not due to additional training compute. \our and the learnable query baseline have similar training FLOPs: for each sampled timestep, both run the same forward pass through the frozen LLM and the denoiser, and both backpropagate through these computations while updating only the conditioning inputs or projector and the denoiser. At inference time, the learnable query baseline computes its condition once, while \our runs both models at every denoising step to condition on the current $\bm{z}_t$. To test whether the gain can be explained by additional inference compute, we make the learnable queries timestep-dependent and rerun their encoder at every denoising step, closely matching \our in inference FLOPs. This variant reaches only 0.54 on GenEval, below the original learnable query baseline. This comparison supports the value of an input-dependent conditioning signal over evolving noisy representations.

\subsection{Multimodal Perception Pretraining}
\label{sec:perception_pretraining}
The gains above depend on the conditional encoder being able to interpret structured visual representations along a denoising trajectory. We therefore examine what makes an LLM better at interpreting noisy RAE latents.

In particular, our conditional encoder is an MLLM whose backbone has been pretrained to perceive visual representations. We therefore study the role of multimodal perception pretraining. To isolate the effect of this capability, we compare a language-only LLM backbone with a perception-pretrained MLLM backbone, while keeping the denoiser and token budget the same. As shown in Figure~\ref{fig:ptlm}, replacing the language-only LLM with a perception-pretrained MLLM improves both Transfusion-RAE and \our under settings with frozen and trainable LLMs. This indicates that perception pretraining provides a transferable prior for diffusion in RAE space: an MLLM that can interpret clean visual representations also better supports encoding their noisy counterparts.

We further compare preserving a perception-pretrained LLM with jointly optimizing the LLM for generation. Following Transfusion~\citep{transfusion}, when fine-tuning the LLM, we add an auxiliary language modeling (LM) loss on the caption tokens and allow the injected noisy visual tokens to use bidirectional attention, while keeping the caption stream causal. Figure~\ref{fig:frozenllm} shows a consistent pattern: fine-tuning helps when starting from a language-only LLM backbone (in both VAE and RAE setups), but it degrades performance when starting from a perception-pretrained MLLM backbone in \our-RAE. This suggests that multimodal perception pretraining is a strong prior that is best preserved rather than further re-optimized for generation.

\begin{figure}[!t]
    \centering
    \begin{subfigure}[t]{0.49\linewidth}
        \centering
        \includegraphics[width=\linewidth]{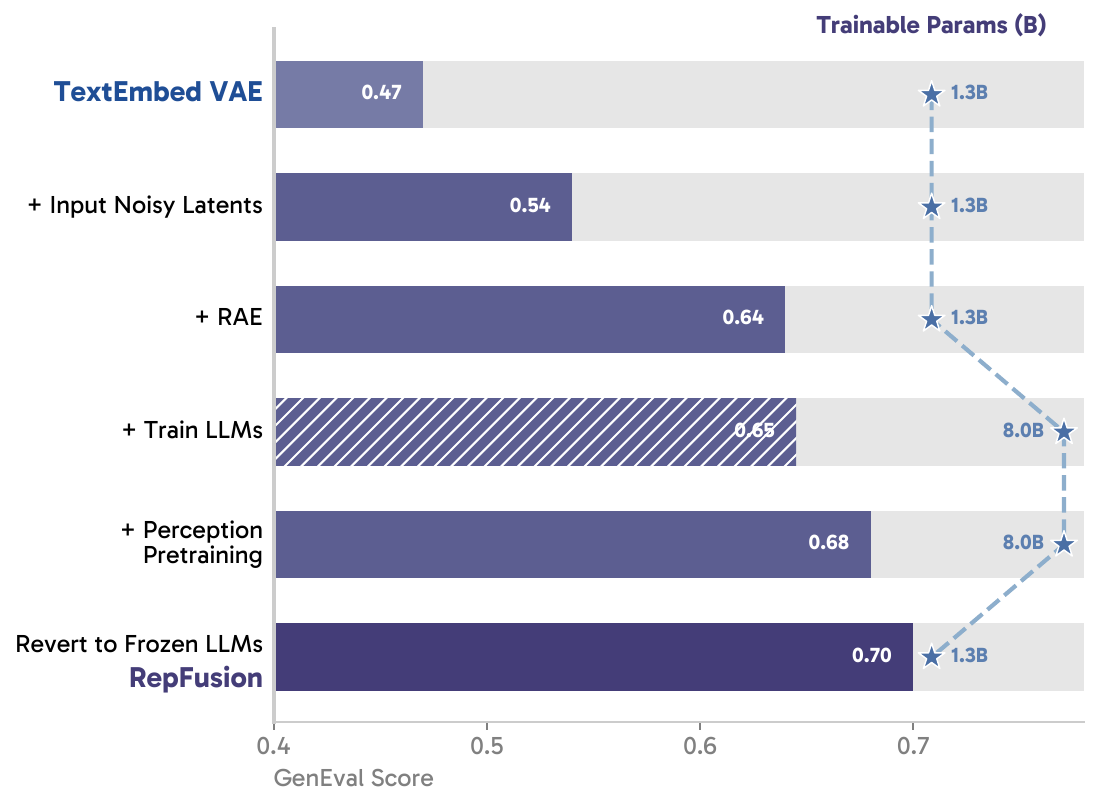}
        \caption{Path from TextEmbed to \our.}
        \label{fig:convnext}
    \end{subfigure}
    \hfill
    \begin{subfigure}[t]{0.49\linewidth}
        \centering
        \includegraphics[width=\linewidth]{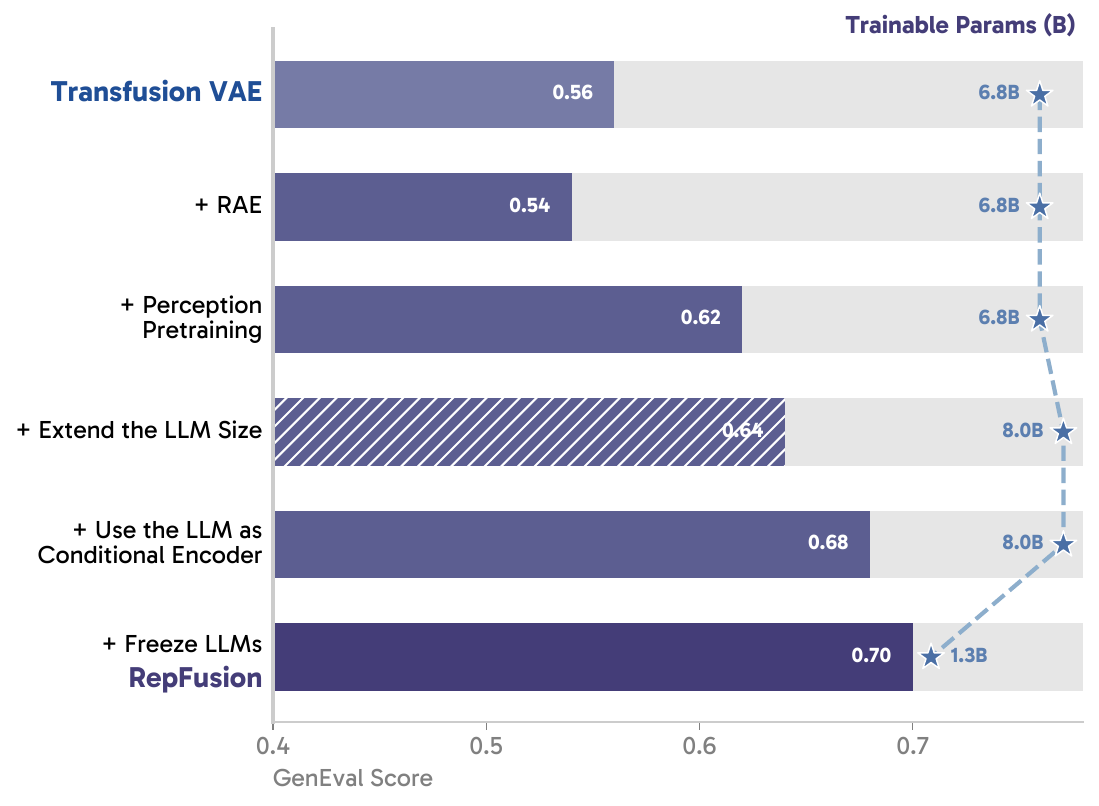}
        \caption{Path from Transfusion to \our.}
        \label{fig:convnext_transfusion}
    \end{subfigure}
    \caption{Step-by-step ablations from (a) TextEmbed and (b) Transfusion toward \our. Bars show GenEval scores, and hatched bars denote modifications that are evaluated but not adopted.}
    \label{fig:convnext_both}
\end{figure}

\subsection{Breaking down the improvement}
\label{sec:breakdown}
We use these observations to break down the improvements of \our over TextEmbed and Transfusion.

As shown in Figure~\ref{fig:convnext}, starting from a standard text embedding baseline with a GenEval score of 0.47, feeding noisy VAE latents into the LLM improves the score to 0.54. Replacing VAE latents with RAE latents, which are easier to denoise and more compatible with LLMs, further improves the score to 0.64. Jointly training the LLM with an LM loss and bidirectional attention gives a minor gain to 0.65. Finally, adding multimodal perception pretraining improves denoising in RAE space, but the best performance is achieved when the LLM backbone remains frozen, preserving the pretrained prior.

We also trace the path from the Transfusion baseline. Transfusion can be viewed as another way of exposing noisy visual latents to language models, but this unified baseline can be improved by using the LLM explicitly as a conditional encoder rather than as the denoiser itself. As shown in Figure~\ref{fig:convnext_transfusion}, starting from the Transfusion baseline at 0.56, replacing VAE latents with RAE latents and adopting a perception-pretrained LLM improves performance to 0.62. We then replicate the last 6 layers of the LLM to construct a stronger 8.0B Transfusion-RAE baseline. Reallocating the same 1.3B trainable parameters to a separate DiT, while using the LLM as the conditional encoder, improves performance from 0.64 to 0.68; freezing the LLM further increases it to 0.70.

\section{Experiments}
\subsection{Experimental Setup}
\label{sec:experimental_setup}

\thinparagraph{Model}
Unless otherwise specified, we follow the MLLM setup of~\citet{llava}: a causal LLM backbone is paired with a CLIP-L/14 vision tower~\citep{clip} through an MLP projector, which provides a clean interface for our RAE setup. We follow this simple architecture because many recent MLLMs introduce fine-tuned vision towers, any-resolution support, token compression~\citep{gemma3}, and deep stacks~\citep{qwen3vl}, which are tailored for multimodal understanding and are non-trivial to adopt for denoising purposes. We set the input resolution to 336, producing $N{=}576$ visual tokens. For VAE-based experiments, we use DC-AE~\citep{dcae} with a spatial downsampling factor of 32. We set the input resolution to 512, which yields $N{=}256$ latent tokens. This setup keeps output resolutions comparable across different latent spaces; we include a token-matched DC-AE comparison in Appendix~\ref{app:token_matching}. For both RAE and VAE settings, we set the DiT patch size to 1.

\thinparagraph{Data}
We pretrain all models on the BLIP-3o 31M dataset~\citep{blip3o}, which is recaptioned with MLLMs and contains 27M long- and 4M short-caption pairs. For supervised fine-tuning (SFT), we combine BLIP-3o 60k~\citep{blip3o}, ShareGPT4o-Image~\citep{sharegpt4oimage}, and Echo-4o~\citep{echo4o} into a 200k synthetic dataset. SFT images are sourced from GPT-4o Image~\citep{gpt4oimagegeneration}.

\thinparagraph{Training}
For all pretraining experiments, we train the models on 128 H200 GPUs with a global batch size of 2{,}048. Models are trained for 10 epochs (160k steps) with a learning rate of $3 \times 10^{-4}$. They are optimized using the AdamW~\citep{adamw} optimizer with $\beta_1 = 0.9$, $\beta_2 = 0.95$, and a weight decay of 0.1. The learning rate follows a cosine decay schedule with a 10k-step warmup period. For SFT experiments, we use a learning rate of $1 \times 10^{-4}$ and train the model for 64 epochs.

\begin{figure*}[!t]
    \centering
    \includegraphics[width=\linewidth]{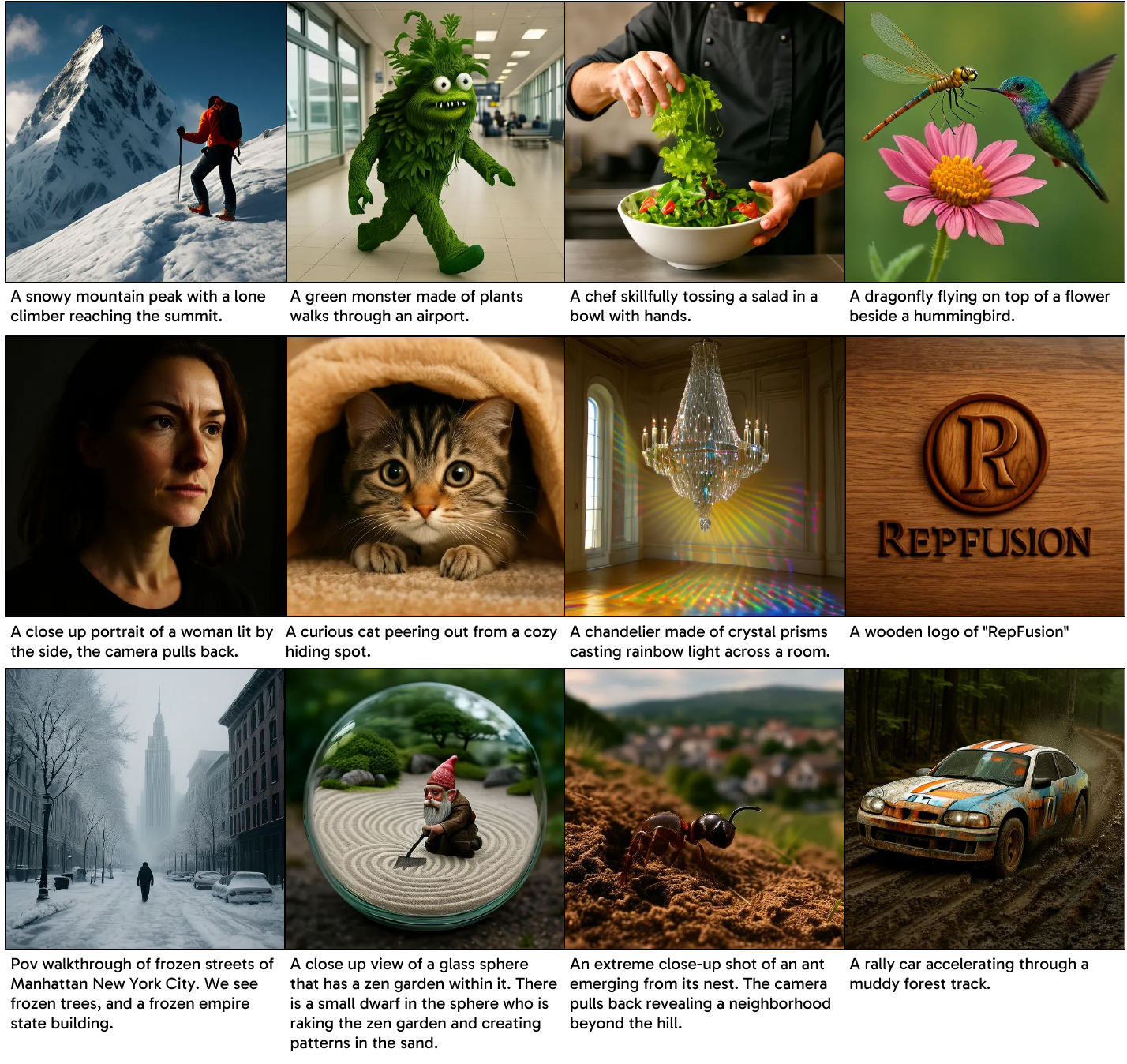}
    \caption{Qualitative T2I samples generated by \our. Some prompts are adapted from~\citet{moviegen}.}
    \label{fig:t2i}
\end{figure*}

\thinparagraph{Representation Decoders}
To decode the visual representations back into pixels, we employ two different strategies: an RAE decoder~\citep{rae} and a diffusion decoder~\citep{emu}. Unless otherwise specified, all parameter counts reported exclude decoder parameters, and the decoder is the RAE decoder by default. For the RAE decoder, we follow standard RAE practice by using a ViT-XL~\citep{mae} decoder and a DINO~\citep{dino} GAN discriminator. The decoder patch size is set to 24 to produce images at a resolution of 576. We train the RAE decoder on ImageNet-22k~\citep{imagenet} for 16 epochs. For the diffusion decoder, we follow Emu~\citep{emu}, starting from the SANA 1.6B checkpoint and replacing the text conditioning with CLIP features. The output resolution is 512. We train the diffusion decoder on ImageNet-22k for 10 epochs.

\begin{table*}[!t]
    \small
    \centering
    \begin{tabular}{lcccc}
        \textbf{Methods} & \textbf{GenEval $\uparrow$} & \textbf{GenEval++ $\uparrow$} & \textbf{GenEval2 $\uparrow$} & \textbf{DPG-Bench $\uparrow$} \\
        \toprule
        Transfusion~\citep{transfusion} & 0.63 & - & - & - \\
        MetaQuery-XL~\citep{metaquery} & 0.80$^\dagger$ & - & - & 82.05 \\
        BLIP-3o 8B~\citep{blip3o} & 0.84 & 0.307 & - & 81.60 \\
        OmniGen2~\citep{omnigen2} & 0.80 & 0.325 & - & 83.57 \\
        BAGEL~\citep{bagel} & 0.82 & 0.371 & 23.1$^\dagger$ & 84.03 \\
        Scale-RAE~\citep{scalerae} & 0.83 & - & - & 79.70 \\
        \midrule
        \our w/ RAE Decoder & 0.73 & 0.432 & 30.2 & 82.75 \\
        \our w/ Diffusion Decoder & 0.78 & 0.443 & 29.9 & 84.41 \\
        \our-SFT w/ RAE Decoder & 0.85 & 0.707 & 35.1 & 84.17 \\
        \our-SFT w/ Diffusion Decoder & 0.87 & 0.669 & 34.9 & 85.11 \\
    \end{tabular}
    \caption{T2I generation results on GenEval~\citep{geneval}, GenEval++~\citep{echo4o}, GenEval2~\citep{geneval2}, and DPG-Bench~\citep{dpg}. $^\dagger$ denotes rewritten prompts. For GenEval2, we report the prompt-level metric Soft-TIFA\textsubscript{GM}.}
    \label{tab:results}
\end{table*}

\subsection{Prompt Alignment}
\label{sec:prompt_alignment}
With only around 30M image-caption pairs, \our achieves strong T2I prompt alignment, with qualitative samples shown in Figure~\ref{fig:t2i}. We evaluate our largest configuration, which uses a 7B MLLM and a 3.2B DiT, on four representative benchmarks: GenEval~\citep{geneval}, GenEval++~\citep{echo4o}, GenEval2~\citep{geneval2}, and DPG-Bench~\citep{dpg}. As shown in Table~\ref{tab:results}, \our achieves competitive performance across all of them, and \our-SFT further improves the performance to state-of-the-art levels. Notably, benchmarks such as GenEval and DPG-Bench are increasingly subject to benchmark-specific optimization in the current synthetic data era~\citep{reca}: many pipelines perform SFT on synthetic images sourced from GPT-4o~\citep{gpt4oimagegeneration} and Nano Banana~\citep{nanobanana}, or directly apply RL with GenEval as a verifiable reward. To address this benchmark drift issue, GenEval2 was recently proposed with a more robust evaluation protocol, Soft-TIFA. Consistent with this motivation, we find that \our-SFT yields only limited improvements on GenEval2, while the pretrained \our remains strong. \our also compares favorably with BAGEL~\citep{bagel} with Self-CoT, which is pretrained on over 1 billion web-scale examples.

\subsection{Reasoning-based Generation}
\label{sec:reasoning}
We find that, similar to learnable query methods such as MetaQuery~\citep{metaquery} and BLIP-3o~\citep{blip3o}, \our can also effectively leverage the capabilities of a frozen LLM. This enables the model to better understand and follow complex prompts, including those requiring world knowledge and reasoning. To quantitatively evaluate \our's world-knowledge reasoning capability, we employ the WISE~\citep{wise} benchmark. As shown in Table~\ref{tab:wise}, \our matches state-of-the-art performance.

\begin{table}[!t]
    \small
    \centering
    \begin{tabular}{lccccccc}
        \textbf{Model} & \textbf{Cultural} & \textbf{Time} & \textbf{Space} & \textbf{Biology} & \textbf{Physics} & \textbf{Chemistry} & \textbf{Overall} \\
        \toprule
        MetaQuery-XL~\citep{metaquery} & 0.56 & 0.55 & 0.62 & 0.49 & 0.63 & 0.41 & 0.55 \\
        BLIP-3o 8B~\citep{blip3o} &-&-&-&-&-&-& 0.62 \\
        BAGEL~\citep{bagel} & 0.44 & 0.55 & 0.68 & 0.44 & 0.60 & 0.39 & 0.52 \\
        \midrule
        \our-SFT w/ RAE Decoder & 0.55 & 0.53 & 0.70 & 0.51 & 0.57 & 0.41 & 0.55 \\
        \our-SFT w/ Diffusion Decoder & 0.65 & 0.63 & 0.79 & 0.63 & 0.67 & 0.44 & 0.64 \\
    \end{tabular}
    \caption{Reasoning-based generation on WISE~\citep{wise}.}
    \label{tab:wise}
\end{table}

\subsection{Conditioning Interface}
We ablate how the MLLM hidden states are injected into the DiT. Cross attention provides a general conditioning mechanism, but it adds extra attention projections and treats the conditioning stream as a separate context. In our RAE setting, the $N$ MLLM outputs are naturally aligned with the $N$ DiT tokens, so a token-wise adaptive normalization interface can use this correspondence directly. As shown in Table~\ref{tab:conditioning_interface}, AdaLN-Single~\citep{pixart} achieves a slightly higher GenEval score with fewer parameters, and we therefore use it as the default conditioning interface.

\begin{table}[!h]
\small
\centering
\begin{tabular}{lcc}
    \textbf{Method} & \textbf{\# Params} & \textbf{GenEval $\uparrow$} \\
    \toprule
    Cross Attention & 1.6B & 0.69 \\
    AdaLN-Single~\citep{pixart} & 1.3B & 0.70 \\
    \end{tabular}
\caption{Ablation on the interface used to inject MLLM hidden states into the DiT. Parameter counts include the DiT and the conditioning interface.}
\label{tab:conditioning_interface}
\end{table}

\subsection{Scaling Behavior}
\label{sec:scaling}
\begin{figure*}[!t]
    \centering
    \includegraphics[width=\linewidth]{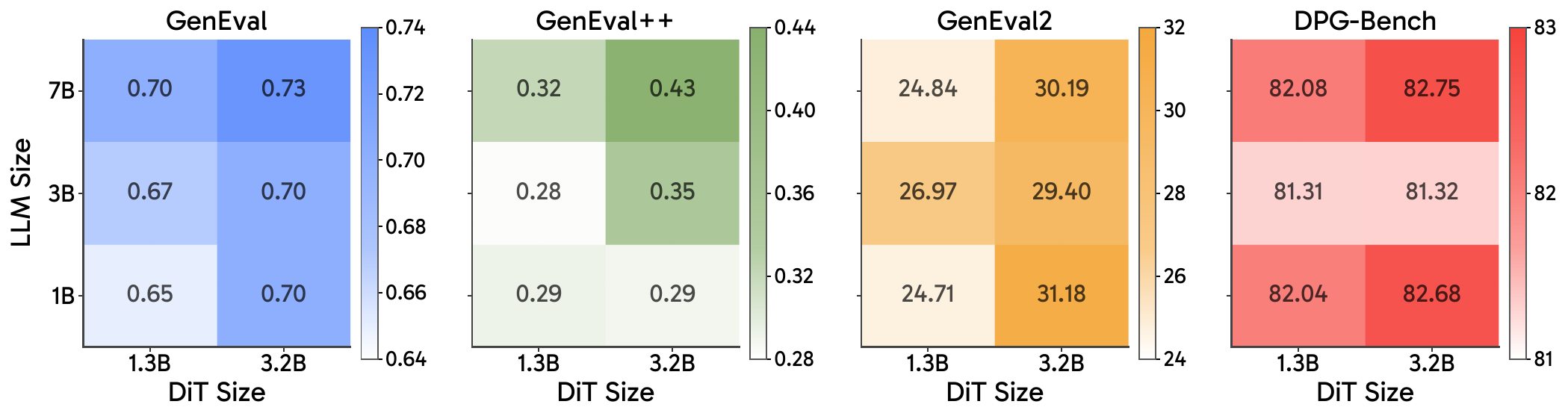}
    \caption{MLLM and DiT co-scaling results. Increasing either component can improve performance, with clearer trends on GenEval and GenEval++.}
    \label{fig:coscaling}
\end{figure*}

\our has two main components: the frozen MLLM that encodes noisy representations, and the DiT denoiser that predicts the velocity. We study how performance changes when scaling either component in the billion-parameter regime, using per-step MLLM recomputation throughout. Figure~\ref{fig:coscaling} shows that increasing either the MLLM or DiT size can improve performance, with the clearest scaling trends appearing on GenEval and GenEval++.

We further study how to allocate inference compute across these two components under iso-FLOPs settings in Table~\ref{tab:iso_flops}. At around 280T FLOPs, allocating more compute to the DiT outperforms the configuration that allocates more compute to the MLLM across all metrics. The same trend holds at around 540T FLOPs: configurations with larger DiTs achieve stronger GenEval++ and GenEval2 scores. These within-family comparisons answer a different question from Figure~\ref{fig:param_efficiency}: among \our variants, scaling the DiT is generally more effective under a fixed inference budget, but compared with TextEmbed, \our remains stronger even when the baseline spends nearly all sampling compute on the DiT. Thus, pretrained MLLM priors complement denoiser capacity, improving generation without eliminating the benefits of scaling the DiT.

\begin{table}[!t]
    \small
    \centering
    \begin{tabular}{ccccccc}
        \textbf{LLM Size} & \textbf{DiT Size} & \textbf{FLOPs Split (LLM $\mid$ DiT)} & \textbf{GenEval $\uparrow$} & \textbf{GenEval++ $\uparrow$} & \textbf{GenEval2 $\uparrow$} & \textbf{DPG-Bench $\uparrow$} \\
        \toprule
        \multicolumn{7}{c}{$\sim$280T inference FLOPs} \\
        \midrule
        1B & 3.2B & \flopsbreakdown{0.20}{0.60}{26\%}{74\%} & 0.70 & 0.289 & 31.18 & 82.68 \\
        3B & 1.3B & \flopsbreakdown{0.59}{0.23}{71\%}{28\%} & 0.67 & 0.282 & 26.97 & 81.31 \\
        \midrule
        \multicolumn{7}{c}{$\sim$540T inference FLOPs} \\
        \midrule
        \rowcolor{black!5}
        \begin{tabular}[c]{@{}c@{}}\textcolor{black!55}{7B}\\[-0.1ex]\scriptsize\strut\end{tabular} & \begin{tabular}[c]{@{}c@{}}\textcolor{black!55}{8.0B}\\[-0.1ex]\makebox[0pt][c]{\hspace{-4.75em}\scriptsize\textcolor{black!55}{(TextEmbed baseline)}}\end{tabular} & \begin{tabular}[c]{@{}c@{}}\flopsbreakdown{0.05}{1.50}{3\%}{97\%}\end{tabular} & \begin{tabular}[c]{@{}c@{}}\textcolor{black!55}{0.66}\end{tabular} & \begin{tabular}[c]{@{}c@{}}\textcolor{black!55}{0.321}\end{tabular} & \begin{tabular}[c]{@{}c@{}}\textcolor{black!55}{26.60}\end{tabular} & \begin{tabular}[c]{@{}c@{}}\textcolor{black!55}{81.34}\end{tabular} \\
        1B & 7.3B & \flopsbreakdown{0.20}{1.35}{13\%}{87\%} & 0.70 & 0.443 & 30.84 & 82.58 \\
        3B & 5.5B & \flopsbreakdown{0.59}{1.00}{37\%}{63\%} & 0.69 & 0.382 & 30.67 & 82.20 \\
        7B & 1.3B & \flopsbreakdown{1.35}{0.24}{85\%}{15\%} & 0.70 & 0.321 & 24.84 & 82.08 \\
    \end{tabular} 
    \caption{Iso-FLOPs comparison. Given a fixed inference budget, we compare different allocations between MLLMs and DiTs within \our, and include TextEmbed as a reference baseline. Scaling the DiT is generally more favorable than scaling the MLLM within \our, but \our remains substantially stronger than TextEmbed, whose static text embedding design leaves nearly all sampling compute in the DiT.}
    \label{tab:iso_flops}
\end{table}

\subsection{MLLM Recomputation}
\label{sec:sparse_updates}
\begin{figure}[!t]
    \centering
    \begin{minipage}{\linewidth}
    \small
    \centering
    \begin{tabular}{ccccc}
        \textbf{MLLM Updates $U$} & \textbf{FLOPs (T)} & \textbf{Latency (s)} & \textbf{Speedup} & \textbf{GenEval $\uparrow$} \\
        \toprule
        30 & 665 & 1.43 & 1.00$\times$ & 0.730 \\
        \midrule
        15 & 440 & 0.97 & 1.48$\times$ & 0.730 \\
        6  & 306 & 0.69 & 2.07$\times$ & 0.725 \\
        3  & 261 & 0.59 & 2.43$\times$ & 0.733 \\
        1  & 231 & 0.52 & 2.73$\times$ & 0.694 \\
    \end{tabular}
    \captionof{table}{Reducing MLLM recomputation during inference. Latency is measured on a single H200 GPU. The $U=30$ row corresponds to the \our w/ RAE Decoder result in Table~\ref{tab:results}. Three MLLM updates achieve performance comparable to per-step MLLM recomputation, with a 2.43$\times$ sampling speedup.}
    \label{tab:sparse_updates}
    \end{minipage}
    \par\medskip
    \includegraphics[width=\linewidth]{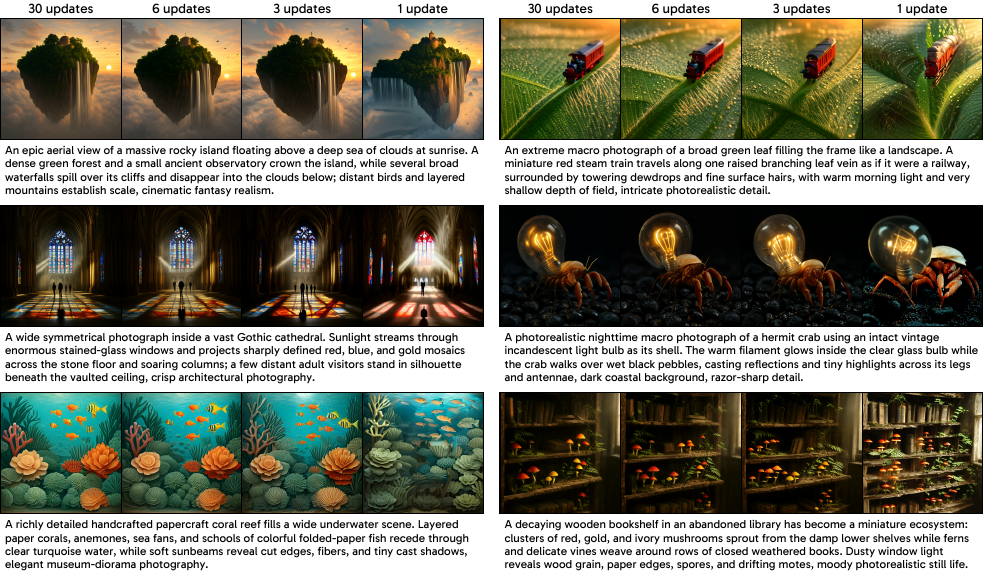}
    \caption{Qualitative comparison of MLLM recomputation. Within each group, the four columns share the same initial-noise seed and use 30, 6, 3, and 1 MLLM updates over 30 denoising steps.}
    \label{fig:interleaved_sampling}
\end{figure}

The \our results above use per-step MLLM recomputation. We now examine whether these pretrained priors can be used with less inference compute. Using \our w/ RAE Decoder from Table~\ref{tab:results}, we reduce the number of MLLM updates $U$ while keeping the trained model and 30 denoising steps unchanged. Table~\ref{tab:sparse_updates} shows that three updates retain performance comparable to per-step recomputation (0.733 vs.\ 0.730 on GenEval), with a 2.43$\times$ sampling speedup. Figure~\ref{fig:interleaved_sampling} shows the qualitative results: the generated images remain visually similar with $U=30$, $6$, and $3$, but change noticeably at $U=1$.

The $U=1$ setting also provides a reference for inference cost: the MLLM runs once and the DiT runs for all 30 denoising steps, as in TextEmbed. For the same model, three updates require only 13\% more inference FLOPs than this single-update setting (261T vs.\ 231T). Together with our controlled comparisons against TextEmbed in Figures~\ref{fig:motivation} and~\ref{fig:param_efficiency}, these results suggest that noisy visual conditioning can provide substantially stronger generation while approaching the inference cost of computing the condition once, as in TextEmbed.

% Keep the recomputation table and qualitative comparison on the same page.
\FloatBarrier

\section{Conclusion}
We study a simple but underused degree of freedom in modern T2I systems: the conditional encoder. By allowing a frozen MLLM to read noisy visual representations, the encoder becomes an active component of the denoising loop rather than a static text encoder. The resulting gains suggest a practical recipe for future models: expose pretrained MLLM priors to evolving noisy representations and preserve those priors. We hope this perspective helps shift how the community thinks about the role of MLLMs in T2I generation.

\clearpage
\newpage
\bibliographystyle{assets/plainnat}
\bibliography{paper}

@String(CVPR= {IEEE Conf. Comput. Vis. Pattern Recog.})

@String(ICCV= {Int. Conf. Comput. Vis.})

@String(ICLR = {Int. Conf. Learn. Represent.})

@String(CVPR  = {CVPR})

@String(ICCV  = {ICCV})

@String(ICLR  = {ICLR})

@misc{flux,
  author={Black Forest Labs},
  title={Flux.1},
  year={2024},
  howpublished={\url{https://bfl.ai/blog/24-08-01-bfl}},
}

@inproceedings{gpt,
  title={Language models are few-shot learners},
  author={Brown, Tom and Mann, Benjamin and Ryder, Nick and Subbiah, Melanie and Kaplan, Jared D and Dhariwal, Prafulla and Neelakantan, Arvind and Shyam, Pranav and Sastry, Girish and Askell, Amanda and others},
  booktitle={NeurIPS},
  year={2020}
}

@inproceedings{transfusion,
  title={Transfusion: Predict the next token and diffuse images with one multi-modal model},
  author={Zhou, Chunting and Yu, Lili and Babu, Arun and Tirumala, Kushal and Yasunaga, Michihiro and Shamis, Leonid and Kahn, Jacob and Ma, Xuezhe and Zettlemoyer, Luke and Levy, Omer},
  booktitle={ICLR},
  year={2025}
}

@inproceedings{luminanext,
  title={Lumina-Next: Making Lumina-T2X Stronger and Faster with Next-DiT},
  author={Zhuo, Le and Du, Ruoyi and Xiao, Han and Li, Yangguang and Liu, Dongyang and Huang, Rongjie and Liu, Wenze and Zhao, Lirui and Wang, Fu-Yun and Ma, Zhanyu and others},
  booktitle={NeurIPS},
  year={2024}
}

@inproceedings{sd3,
  title={Scaling rectified flow transformers for high-resolution image synthesis},
  author={Esser, Patrick and Kulal, Sumith and Blattmann, Andreas and Entezari, Rahim and M{\"u}ller, Jonas and Saini, Harry and Levi, Yam and Lorenz, Dominik and Sauer, Axel and Boesel, Frederic and others},
  booktitle={ICML},
  year={2024}
}

@inproceedings{clip,
  title={Learning transferable visual models from natural language supervision},
  author={Radford, Alec and Kim, Jong Wook and Hallacy, Chris and Ramesh, Aditya and Goh, Gabriel and Agarwal, Sandhini and Sastry, Girish and Askell, Amanda and Mishkin, Pamela and Clark, Jack and others},
  booktitle={ICML},
  year={2021},
}

@inproceedings{llava,
  title={Visual instruction tuning},
  author={Liu, Haotian and Li, Chunyuan and Wu, Qingyang and Lee, Yong Jae},
  booktitle={NeurIPS},
  year={2023}
}

@inproceedings{sd1p5,
  title={High-resolution image synthesis with latent diffusion models},
  author={Rombach, Robin and Blattmann, Andreas and Lorenz, Dominik and Esser, Patrick and Ommer, Bj{\"o}rn},
  booktitle={CVPR},
  year={2022}
}

@inproceedings{vae,
  title={Auto-encoding variational bayes},
  author={Kingma, Diederik P},
  booktitle={ICLR},
  year={2014}
}

@inproceedings{adamw,
  author = {Ilya Loshchilov and Frank Hutter},
  title = {Decoupled Weight Decay Regularization},
  booktitle = {ICLR},
  year = {2019}
}

@inproceedings{imagen,
  author = {Saharia, Chitwan and Chan, William and Saxena, Saurabh and Li, Lala and Whang, Jay and Denton, Emily and Ghasemipour, Seyed Kamyar Seyed and Ayan, Burcu Karagol and Mahdavi, S Sara and Lopes, Rapha Gontijo and others},
  title = {Photorealistic Text-to-Image Diffusion Models with Deep Language Understanding},
  booktitle={NeurIPS},
  year = {2022}
}

@inproceedings{emu,
  author = {Sun, Quan and Yu, Qiying and Cui, Yufeng and Zhang, Fan and Zhang, Xiaosong and Wang, Yueze and Gao, Hongcheng and Liu, Jingjing and Huang, Tiejun and Wang, Xinlong},
  title = {Generative pretraining in multimodality},
  booktitle = {ICLR},
  year = {2024}
}

@inproceedings{pixart,
  title={Pixart-alpha: Fast training of diffusion transformer for photorealistic text-to-image synthesis},
  author={Chen, Junsong and Yu, Jincheng and Ge, Chongjian and Yao, Lewei and Xie, Enze and Wu, Yue and Wang, Zhongdao and Kwok, James and Luo, Ping and Lu, Huchuan and others},
  booktitle={ICLR},
  year={2024}
}

@inproceedings{lmfusion,
  title={LMFusion: Adapting Pretrained Language Models for Multimodal Generation},
  author={Shi, Weijia and Han, Xiaochuang and Zhou, Chunting and Liang, Weixin and Lin, Xi Victoria and Zettlemoyer, Luke and Yu, Lili},
  booktitle = {NeurIPS},
  year={2025}
}

@inproceedings{ddpm,
  title={Denoising diffusion probabilistic models},
  author={Ho, Jonathan and Jain, Ajay and Abbeel, Pieter},
  booktitle={NeurIPS},
  year={2020}
}

@misc{gpt4oimagegeneration,
  title={Introducing 4o Image Generation},
  url={https://openai.com/index/introducing-4o-image-generation/},
  author={OpenAI},
  year={2025}
}

@misc{mistralsmall3,
  title={Mistral Small 3},
  url={https://mistral.ai/news/mistral-small-3},
  author={Mistral AI Team},
  year={2025}
}

@misc{nanobanana,
  title={Introducing Gemini 2.5 Flash Image, our state-of-the-art image model},
  url={https://developers.googleblog.com/introducing-gemini-2-5-flash-image/
},
  author={Google},
  year={2025}
}

@inproceedings{sana,
  title={Sana: Efficient high-resolution image synthesis with linear diffusion transformers},
  author={Xie, Enze and Chen, Junsong and Chen, Junyu and Cai, Han and Tang, Haotian and Lin, Yujun and Zhang, Zhekai and Li, Muyang and Zhu, Ligeng and Lu, Yao and others},
  booktitle={ICLR},
  year={2025}
}

@inproceedings{geneval,
  title={Geneval: An object-focused framework for evaluating text-to-image alignment},
  author={Ghosh, Dhruba and Hajishirzi, Hannaneh and Schmidt, Ludwig},
  booktitle={NeurIPS},
  year={2023}
}

@article{dpg,
  title={Ella: Equip diffusion models with llm for enhanced semantic alignment},
  author={Hu, Xiwei and Wang, Rui and Fang, Yixiao and Fu, Bin and Cheng, Pei and Yu, Gang},
  journal={arXiv preprint arXiv:2403.05135},
  year={2024}
}

@article{llama,
  title={Llama: Open and efficient foundation language models},
  author={Touvron, Hugo and Lavril, Thibaut and Izacard, Gautier and Martinet, Xavier and Lachaux, Marie-Anne and Lacroix, Timoth{\'e}e and Rozi{\`e}re, Baptiste and Goyal, Naman and Hambro, Eric and Azhar, Faisal and others},
  journal={arXiv preprint arXiv:2302.13971},
  year={2023}
}

@article{llama3,
  title={The llama 3 herd of models},
  author={Grattafiori, Aaron and Dubey, Abhimanyu and Jauhri, Abhinav and Pandey, Abhinav and Kadian, Abhishek and Al-Dahle, Ahmad and Letman, Aiesha and Mathur, Akhil and Schelten, Alan and Vaughan, Alex and others},
  journal={arXiv preprint arXiv:2407.21783},
  year={2024}
}

@inproceedings{imagenet,
  title={Imagenet: A large-scale hierarchical image database},
  author={Deng, Jia and Dong, Wei and Socher, Richard and Li, Li-Jia and Li, Kai and Fei-Fei, Li},
  booktitle={CVPR},
  year={2009},
}

@article{moviegen,
  title={Movie Gen: A Cast of Media Foundation Models},
  author={Polyak, Adam and Zohar, Amit and Brown, Andrew and Tjandra, Andros and Sinha, Animesh and Lee, Ann and Vyas, Apoorv and Shi, Bowen and Ma, Chih-Yao and Chuang, Ching-Yao and others},
  journal={arXiv preprint arXiv:2410.13720},
  year={2024}
}

@inproceedings{wise,
  title={WISE: A World Knowledge-Informed Semantic Evaluation for Text-to-Image Generation},
  author={Niu, Yuwei and Ning, Munan and Zheng, Mengren and Lin, Bin and Jin, Peng and Liao, Jiaqi and Ning, Kunpeng and Zhu, Bin and Yuan, Li},
  booktitle={ICML},
  year={2026}
}

@article{metaquery,
  title={Transfer between modalities with metaqueries},
  author={Pan, Xichen and Shukla, Satya Narayan and Singh, Aashu and Zhao, Zhuokai and Mishra, Shlok Kumar and Wang, Jialiang and Xu, Zhiyang and Chen, Jiuhai and Li, Kunpeng and Juefei-Xu, Felix and others},
  journal={arXiv preprint arXiv:2504.06256},
  year={2025}
}

@inproceedings{lipman2022flow,
  title={Flow matching for generative modeling},
  author={Lipman, Yaron and Chen, Ricky TQ and Ben-Hamu, Heli and Nickel, Maximilian and Le, Matt},
  booktitle={ICLR},
  year={2023}
}

@inproceedings{albergo2022building,
  title={Building normalizing flows with stochastic interpolants},
  author={Albergo, Michael S and Vanden-Eijnden, Eric},
  booktitle={ICLR},
  year={2023}
}

@inproceedings{liu2022flow,
    title={Flow Straight and Fast: Learning to Generate and Transfer Data with Rectified Flow},
    author={Xingchao Liu and Chengyue Gong and Qiang Liu},
    booktitle={ICLR},
    year={2023},
}

@inproceedings{rae,
  title={Diffusion transformers with representation autoencoders},
  author={Zheng, Boyang and Ma, Nanye and Tong, Shengbang and Xie, Saining},
  booktitle={ICLR},
  year={2026}
}

@article{t5,
  title={Exploring the limits of transfer learning with a unified text-to-text transformer},
  author={Raffel, Colin and Shazeer, Noam and Roberts, Adam and Lee, Katherine and Narang, Sharan and Matena, Michael and Zhou, Yanqi and Li, Wei and Liu, Peter J},
  journal={JMLR},
  year={2020}
}

@article{blip3o,
  title={Blip3-o: A family of fully open unified multimodal models-architecture, training and dataset},
  author={Chen, Jiuhai and Xu, Zhiyang and Pan, Xichen and Hu, Yushi and Qin, Can and Goldstein, Tom and Huang, Lifu and Zhou, Tianyi and Xie, Saining and Savarese, Silvio and others},
  journal={arXiv preprint arXiv:2505.09568},
  year={2025}
}

@article{sharegpt4oimage,
  title={ShareGPT-4o-Image: Aligning Multimodal Models with GPT-4o-Level Image Generation},
  author={Chen, Junying and Cai, Zhenyang and Chen, Pengcheng and Chen, Shunian and Ji, Ke and Wang, Xidong and Yang, Yunjin and Wang, Benyou},
  journal={arXiv preprint arXiv:2506.18095},
  year={2025}
}

@article{echo4o,
  title={Echo-4o: Harnessing the power of gpt-4o synthetic images for improved image generation},
  author={Ye, Junyan and Jiang, Dongzhi and Wang, Zihao and Zhu, Leqi and Hu, Zhenghao and Huang, Zilong and He, Jun and Yan, Zhiyuan and Yu, Jinghua and Li, Hongsheng and others},
  journal={arXiv preprint arXiv:2508.09987},
  year={2025}
}

@inproceedings{mae,
  title={Masked autoencoders are scalable vision learners},
  author={He, Kaiming and Chen, Xinlei and Xie, Saining and Li, Yanghao and Doll{\'a}r, Piotr and Girshick, Ross},
  booktitle={CVPR},
  year={2022}
}

@inproceedings{dino,
  title={Emerging properties in self-supervised vision transformers},
  author={Caron, Mathilde and Touvron, Hugo and Misra, Ishan and J{\'e}gou, Herv{\'e} and Mairal, Julien and Bojanowski, Piotr and Joulin, Armand},
  booktitle={ICCV},
  year={2021}
}

@inproceedings{dit,
  title={Scalable diffusion models with transformers},
  author={Peebles, William and Xie, Saining},
  booktitle={ICCV},
  year={2023}
}

@inproceedings{attngan,
  title={Attngan: Fine-grained text to image generation with attentional generative adversarial networks},
  author={Xu, Tao and Zhang, Pengchuan and Huang, Qiuyuan and Zhang, Han and Gan, Zhe and Huang, Xiaolei and He, Xiaodong},
  booktitle={CVPR},
  year={2018}
}

@inproceedings{ddt,
  title={Ddt: Decoupled diffusion transformer},
  author={Wang, Shuai and Tian, Zhi and Huang, Weilin and Wang, Limin},
  booktitle={CVPR},
  year={2026}
}

@article{bagel,
  title={Emerging properties in unified multimodal pretraining},
  author={Deng, Chaorui and Zhu, Deyao and Li, Kunchang and Gou, Chenhui and Li, Feng and Wang, Zeyu and Zhong, Shu and Yu, Weihao and Nie, Xiaonan and Song, Ziang and others},
  journal={arXiv preprint arXiv:2505.14683},
  year={2025}
}

@article{qwenimage,
  title={Qwen-image technical report},
  author={Wu, Chenfei and Li, Jiahao and Zhou, Jingren and Lin, Junyang and Gao, Kaiyuan and Yan, Kun and Yin, Sheng-ming and Bai, Shuai and Xu, Xiao and Chen, Yilei and others},
  journal={arXiv preprint arXiv:2508.02324},
  year={2025}
}

@inproceedings{omnigen2,
  title={OmniGen2: Exploration to Advanced Multimodal Generation},
  author={Wu, Chenyuan and Zheng, Pengfei and Yan, Ruiran and Xiao, Shitao and Luo, Xin and Wang, Yueze and Li, Wanli and Jiang, Xiyan and Liu, Yexin and Zhou, Junjie and others},
  booktitle={CVPR},
  year={2026}
}

@inproceedings{scalerae,
  title={Scaling Text-to-Image Diffusion Transformers with Representation Autoencoders},
  author={ Shengbang Tong and Boyang Zheng and Ziteng Wang and Bingda Tang and Nanye Ma and Ellis Brown and Jihan Yang and Rob Fergus and Yann LeCun and Saining Xie },
  booktitle={CVPR},
  year={2026}
}

@inproceedings{ganintcls,
  title={Generative adversarial text to image synthesis},
  author={Reed, Scott and Akata, Zeynep and Yan, Xinchen and Logeswaran, Lajanugen and Schiele, Bernt and Lee, Honglak},
  booktitle={ICML},
  year={2016}
}

@inproceedings{gan,
  title={Generative adversarial nets},
  author={Goodfellow, Ian J and Pouget-Abadie, Jean and Mirza, Mehdi and Xu, Bing and Warde-Farley, David and Ozair, Sherjil and Courville, Aaron and Bengio, Yoshua},
  booktitle={NeurIPS},
  year={2014}
}

@inproceedings{lstm,
  title={Long short-term memory},
  author={Hochreiter, Sepp and Schmidhuber, J{\"u}rgen},
  booktitle={Neural computation},
  year={1997},
}

@misc{flux2,
    author={Black Forest Labs},
    title={{FLUX.2: Frontier Visual Intelligence}},
    year={2025},
    howpublished={\url{https://bfl.ai/blog/flux-2}},
}

@article{zimage,
  title={Z-Image: An Efficient Image Generation Foundation Model with Single-Stream Diffusion Transformer},
  author={Cai, Huanqia and Cao, Sihan and Du, Ruoyi and Gao, Peng and Hoi, Steven and Hou, Zhaohui and Huang, Shijie and Jiang, Dengyang and Jin, Xin and Li, Liangchen and others},
  journal={arXiv preprint arXiv:2511.22699},
  year={2025}
}

@inproceedings{stackgan,
  title={Stackgan: Text to photo-realistic image synthesis with stacked generative adversarial networks},
  author={Zhang, Han and Xu, Tao and Li, Hongsheng and Zhang, Shaoting and Wang, Xiaogang and Huang, Xiaolei and Metaxas, Dimitris N},
  booktitle={ICCV},
  year={2017}
}

@inproceedings{dcae,
  title={Deep compression autoencoder for efficient high-resolution diffusion models},
  author={Chen, Junyu and Cai, Han and Chen, Junsong and Xie, Enze and Yang, Shang and Tang, Haotian and Li, Muyang and Lu, Yao and Han, Song},
  booktitle={ICLR},
  year={2025}
}

@article{gemma3,
  title={Gemma 3 technical report},
  author={Team, Gemma and Kamath, Aishwarya and Ferret, Johan and Pathak, Shreya and Vieillard, Nino and Merhej, Ramona and Perrin, Sarah and Matejovicova, Tatiana and Ram{\'e}, Alexandre and Rivi{\`e}re, Morgane and others},
  journal={arXiv preprint arXiv:2503.19786},
  year={2025}
}

@article{qwen3vl,
  title={Qwen3-VL Technical Report},
  author={Shuai Bai and Yuxuan Cai and Ruizhe Chen and Keqin Chen and Xionghui Chen and Zesen Cheng and Lianghao Deng and Wei Ding and Chang Gao and Chunjiang Ge and Wenbin Ge and Zhifang Guo and Qidong Huang and Jie Huang and Fei Huang and Binyuan Hui and Shutong Jiang and Zhaohai Li and Mingsheng Li and Mei Li and Kaixin Li and Zicheng Lin and Junyang Lin and Xuejing Liu and Jiawei Liu and Chenglong Liu and Yang Liu and Dayiheng Liu and Shixuan Liu and Dunjie Lu and Ruilin Luo and Chenxu Lv and Rui Men and Lingchen Meng and Xuancheng Ren and Xingzhang Ren and Sibo Song and Yuchong Sun and Jun Tang and Jianhong Tu and Jianqiang Wan and Peng Wang and Pengfei Wang and Qiuyue Wang and Yuxuan Wang and Tianbao Xie and Yiheng Xu and Haiyang Xu and Jin Xu and Zhibo Yang and Mingkun Yang and Jianxin Yang and An Yang and Bowen Yu and Fei Zhang and Hang Zhang and Xi Zhang and Bo Zheng and Humen Zhong and Jingren Zhou and Fan Zhou and Jing Zhou and Yuanzhi Zhu and Ke Zhu},
  journal={arXiv preprint arXiv:2511.21631},
  year={2025}
}

@article{geneval2,
  title={GenEval 2: Addressing Benchmark Drift in Text-to-Image Evaluation},
  author={Kamath, Amita and Chang, Kai-Wei and Krishna, Ranjay and Zettlemoyer, Luke and Hu, Yushi and Ghazvininejad, Marjan},
  journal={arXiv preprint arXiv:2512.16853},
  year={2025}
}

@inproceedings{reca,
  title={Reconstruction alignment improves unified multimodal models},
  author={Xie, Ji and Darrell, Trevor and Zettlemoyer, Luke and Wang, XuDong},
  booktitle={ICLR},
  year={2026}
}

\clearpage
\newpage
\beginappendix

\section{Details on Training TextEmbed and Transfusion Baselines}
\label{app:baseline_training}

We train the TextEmbed and Transfusion baselines with the same training setup as \our. The controlled comparisons use the same text encoder family and newly initialized denoising components at the corresponding model scale. For TextEmbed, we follow the recent T2I practice used in Sana~\citep{sana}: the LLM is used as a static text encoder, and its last-layer text-token embeddings condition a newly initialized DiT denoiser. Following Sana, we also apply RMSNorm after the decoder-only text encoder to normalize the variance of the text embeddings to 1.0. For Transfusion~\citep{transfusion}, in addition to the diffusion training described above, we perform interleaved image-captioning training for the same number of training steps, using the same image-caption data.

\section{Impact of Decoders}
\label{app:decoders}
\begin{figure}[!h]
    \centering
    \makebox[\linewidth][r]{%
        \begin{overpic}[width=0.97\linewidth]{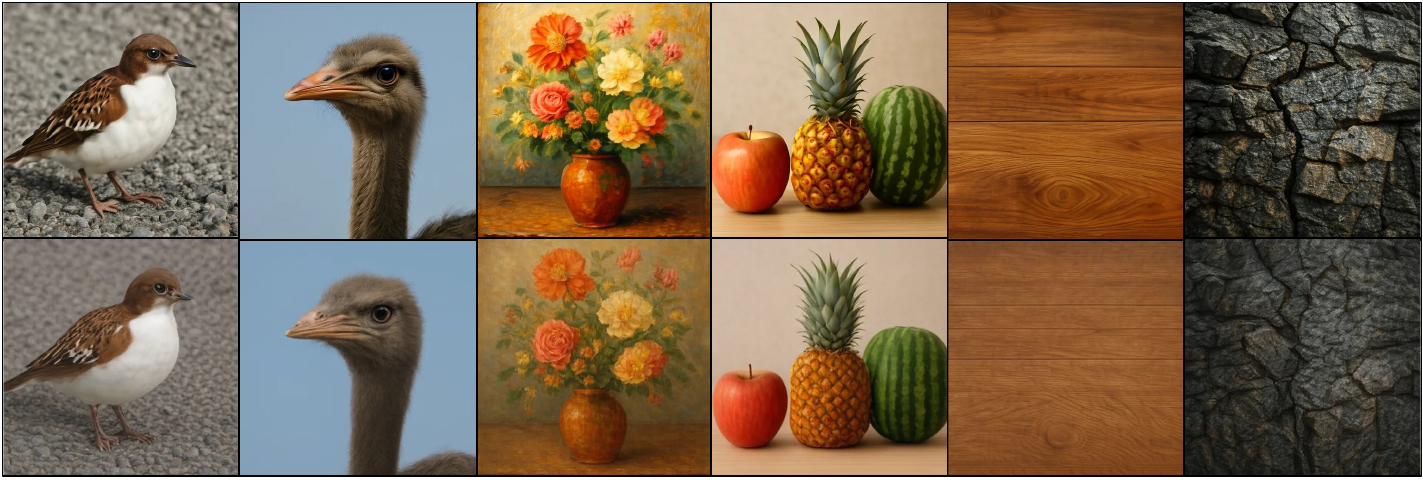}
            \put(-2,20){\rotatebox{90}{\small Diff Decoder}}
            \put(-2,7){\rotatebox{90}{\small RAE}}
        \end{overpic}
    }
\caption{Same CLIP representation decoded by the diffusion decoder and the RAE decoder. Since both decoders are optimized for reconstruction, the mapping from CLIP features to pixels is largely deterministic: object layout and colors are determined upstream when the model denoises the CLIP representation. The choice of decoder mainly affects fine-grained appearance (e.g., textures), and thus has little impact on object-level prompt-following.}
\label{fig:decoder}
\end{figure}

We report \our results with both the RAE decoder and the diffusion decoder in Table~\ref{tab:results}. We observe a performance gap between these two decoders. However, when we use \our to generate a CLIP feature and then apply these two decoders to the same feature, the resulting images appear very similar. The overall layout and colors are largely determined by the CLIP feature, while only fine-grained textures differ slightly (Figure~\ref{fig:decoder}). This suggests that the choice of decoder does not affect the prompt-following ability of \our. Instead, part of the performance gap on GenEval and DPG-Bench appears to arise because images decoded by the RAE decoder are blurrier in texture and therefore harder for detectors or vision-language models (VLMs) to evaluate reliably.

\begin{table}[!h]
    \small
    \centering
    \begin{tabular}{llccc}
        \textbf{Method} & \textbf{Latent Space} & \textbf{Resolution} & \textbf{Sequence Length} & \textbf{GenEval $\uparrow$} \\
        \toprule
        TextEmbed & DC-AE & 768 & 576 & 0.45 \\
        TextEmbed & DC-AE & 512 & 256 & 0.47 \\
        TextEmbed & RAE & 576 & 576 & 0.57 \\
    \end{tabular}
    \caption{Token-matched latent-space comparison for TextEmbed. Increasing the DC-AE sequence length to match the RAE setting does not close the gap to RAE latents.}
    \label{tab:token_matching}
\end{table}

\section{Additional Latent Space Comparison}
\label{app:token_matching}

In the main experiments, we keep the output resolutions comparable across latent spaces, which gives $N{=}256$ tokens for DC-AE and $N{=}576$ tokens for RAE. To isolate the effect of token count, we also evaluate a DC-AE setting with $N{=}576$ tokens by increasing its output resolution. As shown in Table~\ref{tab:token_matching}, matching the DC-AE token count does not improve the TextEmbed baseline.

\end{document}